\documentclass[letterpaper]{article} 
\usepackage{aaai2027}  
\usepackage[hyphens]{url}  
\usepackage{graphicx} 
\usepackage{natbib}  
\usepackage{caption} 
\usepackage{algorithm}
\usepackage{algorithmic}

\usepackage{newfloat}
\usepackage{listings}

\usepackage{epsfig}
\usepackage{amsmath}
\usepackage{amssymb}
\usepackage{cite}
\usepackage{booktabs}
\usepackage{multirow}
\usepackage{makecell}
\usepackage{stmaryrd}
\usepackage{dsfont}
\usepackage{makecell}

\newcommand{\ptitle}[1]{\noindent\textbf{#1}\hspace{5pt}}

\DeclareCaptionStyle{ruled}{labelfont=normalfont,labelsep=colon,strut=off} 
\floatstyle{ruled}
\newfloat{listing}{tb}{lst}{}
\floatname{listing}{Listing}

\usepackage{booktabs}

\usepackage{multirow}
\usepackage{colortbl}
\usepackage{tabularx}
\usepackage{amsmath, bm}
\usepackage{enumitem}

\definecolor{linecolor2}{RGB}{210, 230, 250}
\title{AtlasVLA: Persistent World-Ego State Modeling for \\ Vision-Language-Action Models}
\author{Guiyu Zhao$^{1,2}$, Longteng Guo$^{1,\dagger}$, Yanghong Mei$^{1,2}$, Zilin Zhu$^{1,2}$, Yu Zhang$^{3}$, Bin Cao$^{1,2}$, \\ MingMing Yu$^{4}$, Xingjian He$^{1}$, Jie Jiang$^{1}$, Jing Liu$^{1,2}$}

\affiliations{
    \textsuperscript{\rm 1}Institute of Automation, Chinese Academy of Sciences, Beijing, China\\
    \textsuperscript{\rm 2}University of Chinese Academy of Sciences, Beijing, China\\
    \textsuperscript{\rm 3}Beijing Freedo Technology Co., Ltd.\\
    \textsuperscript{\rm 4}Beihang University\\
}

\begin{document}

\maketitle

\begin{abstract}
While Vision-Language-Action (VLA) models have advanced embodied AI, their fundamentally reactive paradigm severely limits performance in partially observable and long-horizon tasks. When restricted to a single wrist-mounted camera, they inevitably suffer from \emph{perception forgetting} as objects exit the field of view, and temporal \emph{task-progress forgetting} during multi-step execution. To overcome these bottlenecks, we propose AtlasVLA, a novel framework that transitions from direct reactive manipulation to proactive reasoning through a persistent world-ego state. AtlasVLA features a dual-memory architecture: a 4D Persistent World State Memory that lifts transient 2D observations into a globally updated, voxel-hashed spatial state to resolve visual blind spots, and an Ego-Working State Memory that tracks historical ego state and task progress. By conditioning a diffusion transformer (DiT) on this joint World-Ego state, AtlasVLA enables robust spatial reasoning. Extensive evaluations across LIBERO, RLBench, and real-world benchmarks demonstrate that AtlasVLA achieves state-of-the-art performance using \emph{solely} a wrist camera. Remarkably, it decisively outperforms multi-view baselines, yielding absolute success rate improvements of 9.4\% on LIBERO-Long and 17.5\% in real-world long-horizon tasks.
\end{abstract}

\begin{figure*}[ht]
  \centering
  \includegraphics[width=\textwidth]{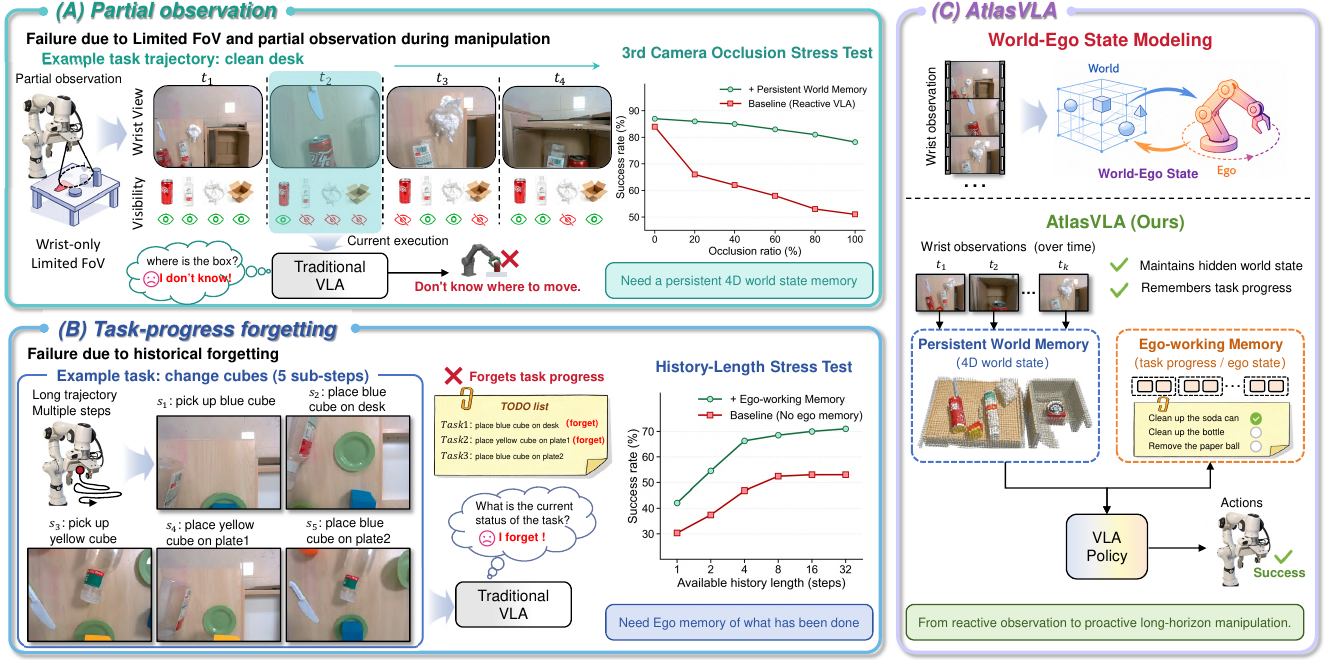}
  \vspace{-6mm}
    \caption{\textbf{The dual bottlenecks of current reactive VLAs versus the advantage of AtlasVLA.} (A) Partial observation: Wrist-only VLAs suffer from limited FoV and partial observation during manipulation. (B) Task-progress forgetting: Reactive baselines lack temporal context, forgetting completed task progress and state. (C) Our solution: Our AtlasVLA integrates a 4D Persistent World State for spatial awareness and an Ego-Working State for task tracking, ensuring robust execution.}
  \label{fig:teaser}
  \vspace{-4mm}
\end{figure*}

\section{Introduction}

Recent advancements in embodied artificial intelligence have witnessed a paradigm shift driven by Vision-Language-Action (VLA) models~\cite{brohan2022rt, zitkovich2023rt, kim2024openvla, black2024pi_0, intelligence2025pi_, shi2025memoryvla}. By mapping multimodal sensory inputs to low-level robotic control commands, these large-scale models~\cite{kim2025fine, black2024pi_0, intelligence2025pi_, shi2025memoryvla} have successfully bridged high-level semantic reasoning and physical execution. However, despite their empirical success, current VLAs operate under a fundamentally reactive paradigm. They predominantly function as reflexive engines that directly map immediate observations to actions, heavily relying on instantaneous visual inputs to dictate physical behavior. 

As illustrated in Figure~\ref{fig:teaser}, this reactive formulation exposes two critical bottlenecks when deployed in unstructured, partially observable environments, particularly under the rigorous solitary wrist-mounted camera setup. First, reactive VLAs suffer from severe \emph{perception forgetting} due to partial observations (Figure~\ref{fig:teaser}(A)). Fundamentally, an instantaneous camera field-of-view (FoV) is not equivalent to the true world state. As the robot manipulates objects, the wrist camera moves dynamically with the end-effector, constantly shifting the FoV. Critical task-relevant objects and spatial structures are immediately forgotten once out of view. Without a persistent internal state, the agent loses track of its spatial surroundings (e.g., "where is the box?"), leading to catastrophic execution failures when operating outside of heavily instrumented multi-view setups.

Second, reactive models are plagued by \emph{task-progress forgetting} during long-horizon manipulation (Figure~\ref{fig:teaser}(B)). Complex tasks, such as sequentially organizing objects or changing configurations, require an agent to execute multiple sub-steps while maintaining a clear cognitive awareness of its progress. Standard VLAs~\cite{intelligence2025pi_,li2024cogact} lack the historical context to remember what they have already accomplished. Without an internal memory of the task progress and ego state, the agent easily loses its place in the overarching execution sequence, causing compounding errors and repeating or missing critical sub-steps.

Unlike current VLAs, human cognition does not rely on an always-visible, omniscient camera to interact with the world. Instead, the human brain seamlessly navigates partial observability by maintaining an internal world model, a continuous cognitive map that tracks objects, structures, and dynamics even when they exit the immediate line of sight. To achieve true autonomy, embodied agents must transcend reactive behaviors and develop the capacity to construct and maintain a \textbf{persistent world-ego state}. This requires transitioning from a paradigm of memoryless observation to one governed by a continuous cycle: \emph{local observation $\rightarrow$ latent state update $\rightarrow$ persistent world state $\rightarrow$ future action}.

To overcome these dual bottlenecks, embodied agents must transcend reactive behaviors and develop the capacity to maintain a persistent hidden state. To this end, we introduce \textbf{AtlasVLA}, a novel {World-Ego State-augmented VLA} framework designed to shift the paradigm from reactive observation to proactive long-horizon manipulation (Figure 1(C)). AtlasVLA continuously reconstructs and updates the hidden state of the environment from limited egocentric observations through a dual-memory architecture.

First, to address partial observability, AtlasVLA employs a \textbf{Persistent World State Memory}. It utilizes streaming depth estimation and spatial back-projection to lift transient 2D wrist observations into a 4D latent space. This continuously updates a global, voxel-hashed persistent world state via neighborhood fusion and sliding windows, enabling the agent to retain the comprehensive workspace observation and effectively overcome the limited FoV of the wrist camera.

Second, to prevent historical forgetting, the persistent world state is integrated with an \textbf{Ego-Working State Memory}. This module establishes an egocentric high-level semantic memory that implicitly represents the historical ego-state and task progress, ensuring the agent constantly remembers what has been done and what needs to be executed next.

Endowed with this world-ego dual-memory architecture, AtlasVLA executes a step-wise conditioned diffusion transformer (DiT) across both persistent world state and ego-working state, generating robust and accurate actions. Extensive experiments demonstrate that AtlasVLA, utilizing \emph{solely} a wrist-mounted camera, achieves state-of-the-art (SOTA) performance ($97.6\%$ success rate on LIBERO and $70.8\%$ on RLBench), decisively outperforming representative multi-view baselines. Notably, compared with $\pi_0$~\cite{black2024pi_0}, AtlasVLA exhibits unprecedented resilience against partial observation and excels in long-horizon memory-dependent tasks, yielding absolute success rate improvements of $9.4\%$ on LIBERO-Long and $17.5\%$ in real-world long-horizon tasks.

Our primary contributions are summarized as follows:
\begin{itemize}
   \setlength{\itemsep}{2pt}
   \setlength{\parsep}{0pt}
   \setlength{\parskip}{0pt}
   \item[$\bullet$] 
   We formally identify the fundamental flaws of current reactive VLAs in wrist-only settings: spatial partial observation and temporal task-progress forgetting.
   \item[$\bullet$] 
   We propose AtlasVLA, a state-augmented VLA framework that seamlessly integrates a 4D Persistent World Memory (to resolve spatial blind spots) and an Ego-Working Memory (to maintain task progress).
   \item[$\bullet$] 
   We design an end-to-end pipeline that transcends the reactive \emph{observe $\rightarrow$ act} paradigm, enabling proactive spatial reasoning from transient local observations.
   \item[$\bullet$] 
   AtlasVLA achieves SOTA performance across LIBERO, RLBench, and real-world tasks using solely a wrist camera, demonstrating exceptional robustness in extreme occlusion and extended-horizon stress tests.
\end{itemize}

\section{Related Work}

\ptitle{Vision-Language-Action Models.} 
The integration of Large Language Models (LLMs)~\cite{touvron2023llama, brown2020language, ouyang2022training} and Vision-Language Models (VLMs)~\cite{GPT4Vision23,llava23,DreamLLM23,ShapeLLM24} has significantly propelled the paradigm shift in embodied artificial intelligence. 
Building upon these foundations, recent VLA models~\cite{zitkovich2023rt, kim2024openvla, kim2025fine, black2024pi_0} have scaled up training by utilizing extensive cross-embodiment datasets (e.g., OXE~\cite{o2024open}, Agibot~\cite{bu2025agibot}) to further enhance generalization. Within this paradigm, foundational models like OpenVLA~\cite{kim2024openvla} formulate continuous control as an autoregressive token generation process. Conversely, recent frameworks such as $\pi_0$~\cite{black2024pi_0}, CogACT~\cite{li2024cogact}, DexVLA~\cite{wen2025dexvla}, DreamVLA~\cite{zhang2025dreamvla} adopt diffusion-based action heads~\cite{chi2025diffusion, liu2025rdt}, leveraging iterative denoising to synthesize complex, multimodal continuous trajectories. Despite these performance gains, these approaches remain heavily dependent on multi-view observation configurations~\cite{fan2026peafowl, black2024pi_0, jangir2022look} or third-person setups~\cite{kim2024openvla, zhang20254d, zhang2025dreamvla, qu2025spatialvla}, fundamentally failing to resolve the wrist-only bottleneck. Furthermore, these methods~\cite{kim2024openvla, brohan2022rt, black2024pi_0} suffer a substantial degradation in performance when confronted with long-horizon tasks.

\ptitle{Memory for Robotic Manipulation.}
Memory has emerged as a critical component for robotic manipulation, particularly in long-horizon tasks. While early approaches implicitly aggregate past observations via recurrent architectures or extended context windows~\cite{brohan2022rt, jiang2023vima, chi2025diffusion}, they struggle with limited capacity and escalating computational costs. To address this, recent works introduce explicit memory mechanisms. For instance,
MAP-VLA~\cite{li2025map} utilizes memory-augmented prompting to deeply contextualize action generation.
MemoryVLA~\cite{shi2025memoryvla} and ReMem-VLA~\cite{li2026remem} leverage retrieval-based visual banks and recurrent latent queries, respectively, to enable efficient long-term temporal reasoning without maintaining full histories. Building upon this, MEM~\cite{torne2026mem} proposes a multi-scale embodied memory architecture to capture hierarchical experiences across varying time horizons. Despite these advancements, existing paradigms remain heavily skewed toward temporal caching without explicit spatial modeling, leaving the inherent partial observability unresolved. Although SOMA~\cite{li2026spatial} attempts to address out-of-vision scenarios using spatial representations, it relies on a static, pre-manipulation snapshot of the instance objects. Devoid of complete scene updating during execution, it remains incapable of supporting continuous wrist-only manipulation or effectively solving extended long-horizon tasks.

\section{Method}

\begin{figure*}[t]
    \centering{\includegraphics[width=1.0\textwidth]{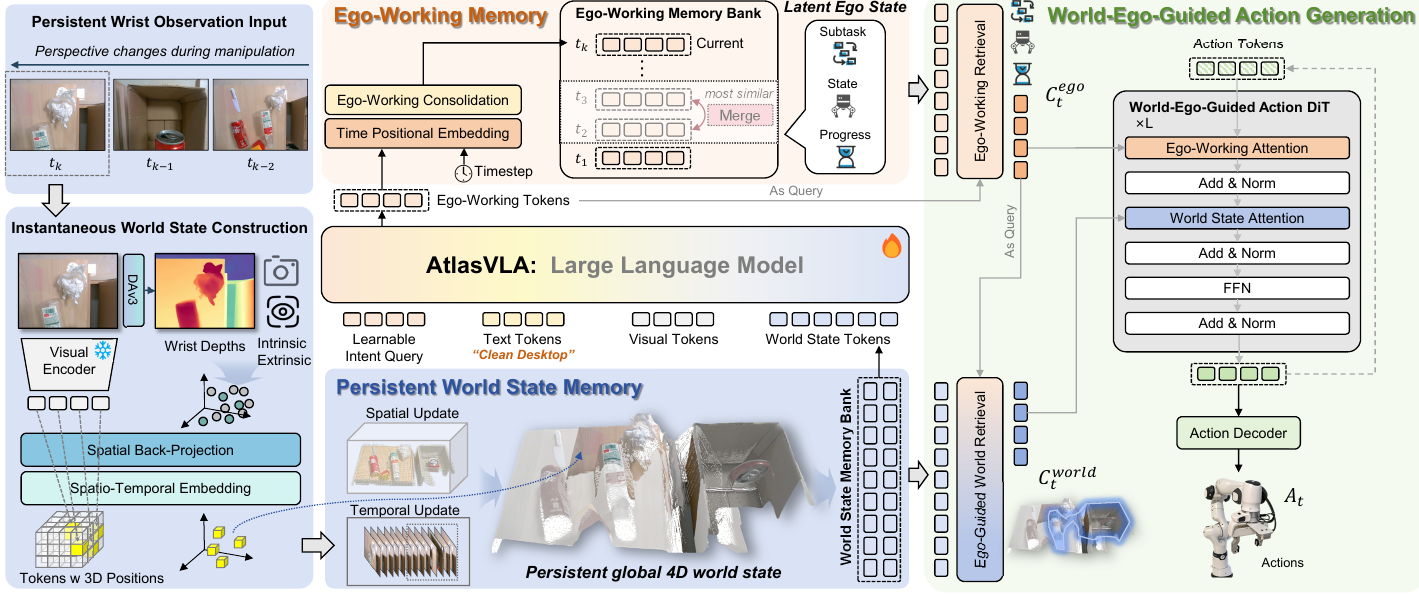}}  
    \vspace{-5mm}
    \caption{\textbf{Overall Architecture of AtlasVLA.} Relying solely on wrist-camera images $I^{w}_t$, AtlasVLA extracts visual tokens $X^{w}_t$ and lifts them into 3D via depth $D^{w}_t$ and extrinsics $\mathbf{T}^{ex}_t$ to form an instantaneous world state memory $m_t$. This memory dynamically updates the persistent world state memory $\mathcal{M}_{t-1}$ via spatial neighborhood fusion and a temporal sliding window. Concurrently, an ego-working memory condenses context into ego-working tokens via learnable queries $Q^{ego}$, updating an ego-working memory bank through memory consolidation. 
    Finally, world state and ego-working retrievals extract contexts $C^{world}_t$ and $C^{ego}_t$ from the dual memory banks to condition the world-ego-guided action DiT, yielding robust actions $A_t$.
    }
\label{pipeline}
\vspace{-3mm}
\end{figure*}

\subsection{Overview of AtlasVLA}

\ptitle{Problem Formulation.}
Following previous work~\cite{kim2024openvla, black2024pi_0}, we formulate VLA-driven robotic manipulation as a sequential decision-making problem. Departing from previous methods~\cite{kim2025fine, black2024pi_0, li2025bridgevla} that rely heavily on third-person or multi-view inputs, we impose a strict wrist-only observation constraint.  Formally, at any given time step $t$, the agent receives solely the wrist-mounted visual observation $O^{w}_t$, the proprioceptive robot state $S_t$, and a language task instruction $L$. The objective is to learn a mapping function $\pi_\theta(\cdot)$ that generates a sequence of future actions:
\begin{equation}
A_t = [a_t, a_{t+1}, \dots, a_{t+k-1}] \sim \pi_\theta(\cdot \mid O^w_t, S_t, L)
\end{equation}
where $A_t$ denotes the action chunk of size $k$. Each action $a_t \in \mathbb{R}^7$ is a 7-dimensional vector comprising the 6-DoF end-effector pose and the binary state of the gripper.
\ptitle{Overview.}
As illustrated in Figure~\ref{pipeline}, AtlasVLA operates exclusively on single-view wrist observations, organizing a persistent world-ego memory into three core modules:
First, \textbf{Persistent World State Memory} (Sec.~\ref{subsec2}) lifts 2D tokens into a 3D latent space via Depth Anything v3~\cite{lin2025depth} and spatial back-projection, injecting spatio-temporal embeddings to form a 4D world representation. It dynamically maintains a global voxel-hashing map via neighborhood fusion, sliding windows, and a permanent first-frame anchor to preserve global world state.
Second, \textbf{Ego-Working Memory} (Sec.~\ref{subsec3}) tracks ego working state and subtask progression via intent-aware queries and redundancy-aware consolidation, dynamically updating an ego-working memory bank to prevent intention drift.
Finally, \textbf{World-Ego-Guided Action Generation} (Sec.~\ref{subsec5}) executes dual-path retrieval to ground the action expert. Guided by these contexts, a step-wise conditioned DiT sequentially applies ego-working attention and world state attention to generate robust actions.















\subsection{Persistent World State Memory}
\label{subsec2}

\ptitle{Instantaneous World State Construction.}
Given the wrist-camera observation $I^{w}_t$ at the current time step $t$, we extract 2D visual tokens utilizing a frozen vision encoder. 
Concurrently, depth is estimated via Depth Anything v3~\cite{lin2025depth}, where we leverage historical frames through a streaming model~\cite{lin2025depth} to enhance temporal depth consistency. 
Furthermore, we actively acquire the camera's intrinsics $\mathbf{T}^{in}$ and extrinsic $\mathbf{T}_t^{ex}$. 
Specifically, by utilizing the current robot state $S_t$ and the hand-eye calibration matrix $\mathbf{T}^{h2e}$, 
we can derive the extrinsic parameters of the wrist camera $\mathbf{T}^{ex} = \psi(S_t) \cdot \mathbf{T}^{h2e}$ where $\psi(\cdot)$ denotes the transformation from the end-effector state to the pose matrix.
Subsequently, we propose a spatial back-projection module to lift the 2D visual tokens into the 3D latent space:
\begin{equation}
    m_t, P_t = \text{Back-Projection}(X_w^t, D^{w}_t, \mathbf{T}^{in}, \mathbf{T}^{ex}_t),
    \label{eq:back_projection}
\end{equation}
where $m_t$ denotes the current local world state which fundamentally comprises 2D latent tokens with 3D positions.

\ptitle{Spatio-Temporal Embedding.}
To empower the world state with precise spatio-temporal awareness and facilitate its subsequent fusion into the global world state memory, we propose a dual spatio-temporal positional embedding mechanism. 
While we align the 2D tokens to the global 3D space, the representations themselves still lack explicit spatial and temporal context modeling, which inevitably leads to spatial aliasing and temporal degradation. To mitigate this issue, we incorporate 3D spatial coordinates and timestamps to augment the tokens through learnable positional encodings:
\begin{equation}
    \widehat{m}_t = m_t + \mathcal{E}_{spatial}(P_{t}) + \mathcal{E}_{temporal}(t),
    \label{eq:st_embedding}
\end{equation}
where $\mathcal{E}_{spatial}(\cdot)$ encodes the 3D position to preserve geometric structures, and $\mathcal{E}_{temporal}(\cdot)$ injects temporal positional encodings to capture the sequential. 
Both of them are parameterized by MLP.
By entangling spatial and temporal cues, the instantaneous world state transitions from an isolated observation pool into a unified 4D representation.

\ptitle{World State Spatio-Temporal Update.}\label{subsec3}
We maintain a persistent world state memory $\mathcal{M}_t$ based on the local world state $m_t$ extracted at each time step $t$. 
As long-horizon manipulation progresses, directly accumulating working memory inevitably leads to severe spatial redundancy and computational bottlenecks. 
To address this, we introduce a localized spatial fusion mechanism and a temporal sliding window to perform memory updates. 

\textbf{Spatially}, we adopt a voxel-hashing strategy analogous to Truncated Signed Distance Function (TSDF) map integration~\cite{newcombe2011kinectfusion}. 
The entire 3D space is partitioned into uniform voxels. 
For incoming tokens corresponding to identical physical regions, we perform a weighted aggregation of their latent features within each local voxel. 
Let $v$ denote the 3D coordinate of a voxel. 
When a newly acquired world state $m_t(v)$ with confidence weight $w_t$ is projected into the global world, the global world memory $\mathcal{M}_t(v)$ and its cumulative weight $\mathcal{W}_t(v)$ are dynamically updated:
\begin{equation}
    \mathcal{M}_t(v) = \frac{\mathcal{W}_{t-1}(v) \mathcal{M}_{t-1}(v) + w_t m_t(v)}{\mathcal{W}_{t-1}(v) + w_t}.
    \label{eq:spatial_feature_update}
\end{equation}
where confidence weight $w_t$ reflects the reliability of the current observation and is directly derived from the depth estimation confidence:
\begin{equation}
    w_t(v)=c_t(v), \; \mathcal{W}_t(v) = \lambda\mathcal{W}_{t-1}(v)+w_t(v).
    \label{eq:depth_confidence}
\end{equation}
Higher-confidence depth observations contribute more to the global memory, while uncertain measurements are suppressed during the aggregation process.

\textbf{Temporally}, the global memory maintains a maximum temporal window size of $W$,  it will be continuously updated and forgotten through a sliding window.
Crucially, since the first frame typically provides an optimal field-of-view and accurately reflects the initial state of the manipulation, we enforce a strict ``permanent initialization'' rule. 
The spatio-temporal memory constructed from the initial frame is permanently anchored to provide a persistent global context. 






\begin{figure*}[t]
    \centering{\includegraphics[width=1.0\textwidth]{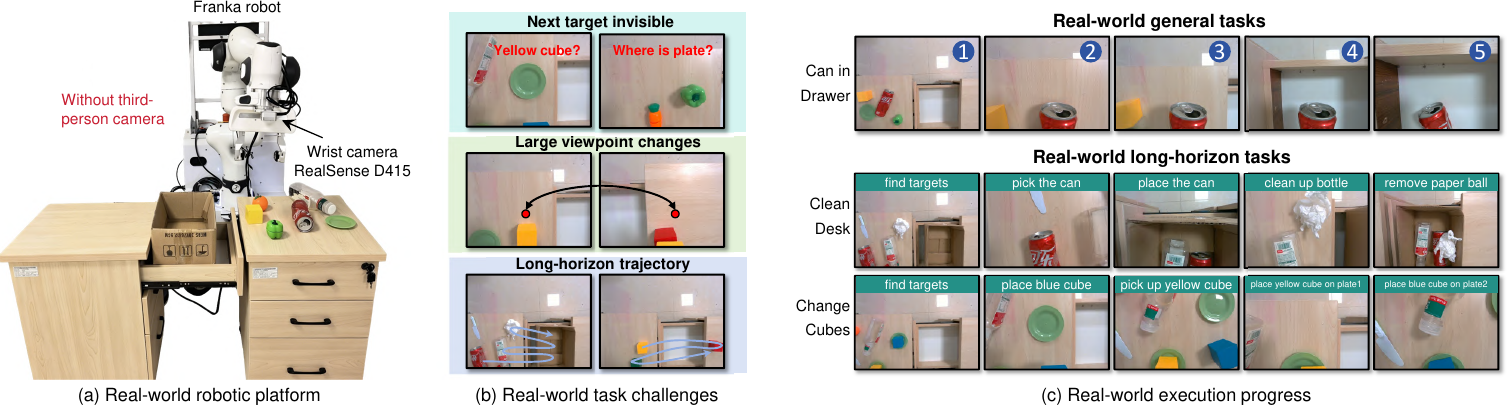}}  
    \vspace{-5mm}
    \caption{
    Qualitative results of AtlasVLA on real-world long-horizon tasks.
    }
    \label{fig:qr_}
 \end{figure*}

  \begin{table*}[htbp]
\setlength{\tabcolsep}{15pt}
\begin{center}
\resizebox{1.0\textwidth}{!}{
\begin{tabular}{l|c|ccccc|c}
\toprule
 \multirow{1}{*}{Method} &Cameras & Spatial & Object & Goal & Long &  90 & \multirow{1}{*}{\textbf{Average}} \\
\midrule
OpenVLA~\scriptsize~\cite{kim2024openvla} &3rd & 84.7& 88.4& 79.2& 53.7& 73.5 & 75.9 \\
$\pi_0$~\scriptsize\citep{black2024pi_0} &3rd &90.8 &91.8 &89.6 &80.2 &-- &88.1\\
4D-VLA~\scriptsize\citep{zhang20254d} &3rd & 93.8 & 92.8 & 95.6 & 86.5 & -- & 92.2 \\
CogACT~\scriptsize\citep{li2024cogact} &3rd & 97.2 & 98.0 & 90.2 & 88.8 & 92.1 & {93.2} \\
MemoryVLA~\scriptsize\citep{shi2025memoryvla} &3rd & \underline{98.4} & 98.4 & 96.4 & 93.4 & \underline{95.6} & {96.5} \\
\midrule
$\pi_0$~\scriptsize\citep{black2024pi_0} &3rd + wrist & 96.8 & 98.8 & 95.8 & 85.2 & -- & 94.2 \\
OpenVLA-OFT &3rd + wrist &97.6 &98.4 &\underline{97.9} &\underline{94.5} &-- &\underline{97.1} \\
GE-ACT~\scriptsize\citep{liao2025genie} &3rd + wrist & 98.2 & 97.6 & 95.8 & 94.4 & -- & 96.5 \\
\midrule
CogACT~\scriptsize~\cite{li2024cogact} &wrist & 96.4 &95.8 &88.6 &86.2  &87.4 &90.9\\
$\pi_0$~\scriptsize\citep{black2024pi_0} &wrist &94.4 &96.6 &90.8 &80.8 &-- &90.7\\
MemoryVLA~\scriptsize\citep{shi2025memoryvla} &wrist &96.2 &\underline{99.2} &96.4 &87.6  &90.7 &94.0\\
{\textbf{AtlasVLA (Ours)}} &wrist & \textbf{99.4}  & \textbf{99.8} & \textbf{98.2} & \textbf{94.6} & \textbf{95.8} &\textbf{97.6}\\
\bottomrule
\end{tabular}}
\vspace{-2mm}
\caption{
\textbf{Quantitative comparison on the LIBERO benchmark.}
Success rates (\%) are averaged over 50 trials per task across five suites, each containing 10 tasks.
For methods without LIBERO-90
results, we report the average over the first four suites.
}
\vspace{-3mm}
\label{tab:libero_main}
\end{center}
\end{table*}

\subsection{Ego-Working State Memory}

\ptitle{Intent-aware Query.}
A critical challenge in long-horizon manipulation is preventing the agent from losing sight of the global objective amidst continuous, low-level sensory streams. 
To address this, we equip the VLA model with an intent-aware query mechanism to egocentrically model and track the task intent (state, progress). 
We define a set of learnable queries as the intent queries, denoted by $Q^{ego} \in \mathbb{R}^{N \times d}$. 
At each decision step, $Q^{ego}$ is fed into the VLM to aggregate goal-oriented information. 
Formally, the queries attend to all the tokens via cross-attention mechanisms:
\begin{equation}
    Z^{ego} = \text{Softmax}\left(\frac{Q^{ego} K^T}{\sqrt{d}}\right) V,
    \label{eq:intent_query}
\end{equation}
where $K$ and $V$ are the key and value projections of the tokens, and $Z^{ego}$ represents the extracted ego-working latent tokens. 
By condensing the extensive spatio-temporal context into these focused intent tokens, our model maintains a coherent global receptive field. 
This explicitly prevents intention forgetting and significantly enhances the agent's capability to execute complex, multi-stage tasks over long horizons.


\ptitle{Ego-Working Memory Bank.}
In long-horizon manipulation, task intentions dynamically evolve across intermediate sub-goals. 
To capture these transitions while avoiding unbounded memory growth and semantic redundancy, we maintain a temporal intent memory bank $\mathcal{M}^{ego}$ through redundancy-aware latent consolidation. 
Given the accumulated memory $\mathcal{M}^{ego}_{t-1}$, the newly extracted ego-working token $Z^{ego}_{t}$ first retrieves relevant historical context, which is adaptively fused into the current representation. 
The memory bank is updated as follows:
\begin{equation}
    \mathcal{M}^{ego}_{t} = \text{Cons}\left( \mathcal{M}^{ego}_{t-1} \cup \{Z^{ego}_t+\mathcal{E}_{temporal}(t)\} \right),
    \label{eq:memory_consolidation}
\end{equation}
where $\text{Cons}(\cdot)$ denotes the consolidation that merges temporally adjacent and semantically similar intent tokens. 

\subsection{World-Ego-Guided Action Generation}
\label{subsec5}


\ptitle{Ego-Working Memory Retrieval.}
To leverage historical context, the current ego-working token $Z^{int}_t$ retrieves relevant information from the consolidated memory $\mathcal{M}^{ego}_{t}$ via cross-attention:
\begin{equation}
    C^{ego}_t = \text{CrossAttn}(Z^{ego}_t, \mathcal{M}^{ego}_{t}, \mathcal{M}^{ego}_{t}).
    \label{eq:intent_retrieval}
\end{equation}
By extracting the most relevant past intents, the retrieved context $C^{ego}_t$ ensures temporally coherent action decoding.

\ptitle{Ego-Guided World Retrieval.}
To ground actions in the 3D environment, the agent retrieves task-relevant world states through an ego-guided world retrieval module. 
Specifically, the historical ego-working context $C^{ego}_t$ serves as the query to attend to the global world memory $\mathcal{M}_t$ via cross-attention:
\begin{equation}
    C^{world}_t = \text{AddNorm}\left(\text{FFN}\left(\text{IntentAttn}(C^{ego}_t, \mathcal{M}_t, \mathcal{M}_t)\right)\right),
    \label{eq:map_retrieval}
\end{equation}
where $\mathcal{M}_t$ provides both keys and values. 
This ego-guided retrieval selectively extracts relevant world states while suppressing spatial redundancy, producing an ego-aligned representation for precise action generation.



\ptitle{World-Ego-Guided Action DiT.}
Unlike standard diffusion models with global conditioning, our DiT adopts a decoupled step-wise conditioning mechanism. 
The action generation is conditioned on both ego-working states for task progression and retrieved world states for geometric grounding.

At each diffusion step $k$, the noisy action tokens $a_k$ are processed by the world-ego-guided action DiT blocks. 
First, the tokens attend to the ego-working context $C^{ego}_t$ via an ego-working attention layer to capture the current ego-working state and task progress. 
Then, they attend to the retrieved world representation $C^{world}_t$ via a world state attention layer to introduce a more comprehensive world observation.
Both attention modules and the subsequent FFN follow the standard Transformer design with Add \& Norm layers. 
After $k$ diffusion steps, the refined tokens are decoded into the final actions $A_t$ by an action decoder.



\begin{table*}[htbp]
\setlength{\tabcolsep}{12pt}
\begin{center}
\resizebox{1.0\textwidth}{!}{
\begin{tabular}{l|c|cccccc|c}
\toprule
\multirow{1}{*}{Method} & Cameras &
\makecell{Sweep to\\Dustpan} &
\makecell{Phone on\\Base} &
\makecell{Umbrella\\Out} &
\makecell{Frame off\\Hanger} &
\makecell{Wine at\\Rack} &
\makecell{Water\\Plants} &
\multirow{1}{*}{\textbf{Avg.}} \\
\midrule
OpenVLA~\scriptsize~\cite{kim2024openvla} & 3rd 
&50.0&20.0&35.0&15.0&10.0&10.0&23.3 \\
CogACT~\scriptsize~\cite{li2024cogact} & 3rd 
&50.0&50.0&55.0&45.0&30.0&25.0&42.5 \\
FiS-VLA~\scriptsize~\cite{chen2026fast} & 3rd 
&\underline{55.0}&50.0&50.0&\textbf{70.0}&55.0&20.0&50.0 \\
MemoryVLA~\scriptsize\citep{shi2025memoryvla} & 3rd 
&50.0&\underline{60.0}&\underline{75.0}&60.0&\textbf{80.0}&\underline{55.0}&\underline{63.3}\\
\midrule
$\pi_0$~\scriptsize\citep{black2024pi_0} &3rd + wrist  
&30.0&30.0&30.0&\textbf{70.0}&10.0&30.0&33.3\\
GE-ACT~\scriptsize~\cite{liao2025genie} &3rd + wrist 
&10.0&15.0&40.0&35.0&40.0&45.0&30.8 \\
\midrule

CogACT~\scriptsize~\cite{li2024cogact} & wrist 
&40.0&35.0&50.0&35.0&20.0&25.0&34.2 \\
MemoryVLA~\scriptsize\citep{shi2025memoryvla} &wrist 
&40.0&55.0&65.0&60.0&60.0&50.0&{55.0}\\

\textbf{AtlasVLA (Ours)} & wrist 
&\textbf{70.0}&\textbf{70.0}&\textbf{80.0}&\underline{65.0}&\underline{75.0}&\textbf{65.0}&\textbf{70.8}\\

\bottomrule
\end{tabular}}
\vspace{-3mm}
\caption{
\textbf{Quantitative comparison on the RLBench benchmark.}
Success rates (\%) are reported on validation over 20 episodes.
}
\vspace{-4mm}
\label{tab:rlbench}
\end{center}
\end{table*}

\section{Experiments}

\subsection{Experimental Setups}

\ptitle{Implementation Details.}
The model is trained across 8 NVIDIA A100 GPUs leveraging PyTorch FSDP, with a per-GPU sample size of 32 (yielding a global batch of 256) and a peak learning rate of $2 \times 10^{-5}$. 
Discarding any third-person views, the network exclusively conditions on a single 224$ \times$ 224 wrist-perspective RGB frame and the text instruction to generate $7\text{-DoF}$ end-effector actions. 
The LLM scales to 7B, while our DiT-based action expert comprises $\sim$300M parameters. 
For online inference, action decoding uses DDIM \cite{song2020denoising} over 10 denoising steps with a classifier-free guidance (CFG) \cite{ho2022classifier} scale of 1.5. Further configurations are detailed in Appendix.

\ptitle{Simulation and Real-world Benchmarks.}
LIBERO~\cite{liu2023libero} uses a Franka robot and spans five distinct suites (Spatial, Object, Goal, Long, and 90). 
RLBench \cite{james2020rlbench} provides a large-scale simulation framework featuring complex, multi-stage manipulation tasks with extensive object and task variations. 
Figure~\ref{fig:qr_} illustrates our comprehensive evaluation across real-world environments, covering the robotic platforms, task challenges, and execution progress.
In the real world, we evaluate our framework on a Franka robot, categorizing experiments into general manipulation and long-horizon temporal suites. 
Crucially, aligning with our core claim, all real-world tasks are conducted using exclusively wrist-camera observations without any third-person visual support. 
Task details for each benchmark and additional qualitative results are provided in Appendix.

\subsection{Simulation Evaluation on LIBERO}

\ptitle{Implementation Setup.}
We evaluate our framework on the LIBERO simulation benchmark \cite{liu2023libero} across five simulation suites: Spatial, Object, Goal, Long, and 90. 
Following~\cite{kim2024openvla, shi2025memoryvla}, models are trained using 50 demonstrations per task. 
Specifically, individual models are trained for the Spatial, Object, and Goal suites for $20{k}$ gradient steps each, whereas a single unified model is trained jointly on Long and LIBERO-90 for $40\text{k}$ steps. 
To demonstrate the robust convergence of our method without validation bias, all reported results are derived directly from the final training checkpoint rather than the best validation step. 
Each task undergoes 50 evaluation rollouts, and the mean success rate across each suite is reported.

\ptitle{Evaluation with Wrist-only Observation.}
We evaluate AtlasVLA under a strict wrist-only camera setting on the LIBERO benchmark. As shown in Table~\ref{tab:libero_main}, previous methods suffer significant degradation when shifting from third-person to wrist-only views, with $\pi_0$ and MemoryVLA dropping by 3.5\% and 2.5\% respectively. Benefiting from our persistent world state memory, AtlasVLA maintains consistently strong performance under this setting, achieving an \textbf{97.6\%} average success rate and even outperforming baselines~\cite{black2024pi_0, shi2025memoryvla, li2024cogact} equipped with additional third-person views, which is \textbf{3.4\%} higher than $\pi_0$. Rather than relying on instantaneous 3rd observations, AtlasVLA continuously accumulates and retrieves spatio-temporal world context from sequential local views, effectively compensating for limited visibility and maintaining coherent scene understanding throughout manipulation.

\ptitle{Evaluation on Long-Horizon Tasks.}
Long-horizon manipulation significantly compounds the challenges of context retention and global scene modeling under wrist-only control. On the demanding LIBERO-Long benchmark, AtlasVLA achieves a state-of-the-art success rate of \textbf{94.6\%} under the strict \textit{wrist-only} constraint, outperforming the strong baseline MemoryVLA by \textbf{7.0\%}. This superiority stems from two core designs: (1) our \textbf{ego-working memory} tracks task progression to prevent intention drift, and (2) our \textbf{persistent world state memory} dynamically builds a comprehensive 3D global workspace. Notably, while MemoryVLA experiences a precipitous \textbf{5.8\%} drop when stripped of third-person views, AtlasVLA's explicit persistent world state modeling effectively eliminates the reliance on external spatial priors.

\subsection{Simulation Evaluation on RLBench}

\ptitle{Implementation Setup.}
We evaluate our framework on RLBench~\cite{james2020rlbench}, training our models for 80k steps using 100 demonstrations per task. Crucially, despite the geometric complexity of RLBench, our agent operates under a strict wrist-only constraint using a single $128 \times 128$ RGB view, discarding all third-person views (front, left and right). Following~\cite{lou2026mask}, we report the average success rate across 6 representative tasks over 20 trials per task directly using the final training checkpoint.

\ptitle{Evaluation Results on RLBench.}
We further evaluate AtlasVLA on RLBench, a challenging testbed featuring multi-stage tasks with larger spatial and trajectory variations than LIBERO. As shown in Table~\ref{tab:rlbench}, AtlasVLA achieves an average success rate of \textbf{70.8\%} using solely a wrist camera, significantly outperforming all VLA baselines. Notably, it surpasses the strongest baseline, MemoryVLA, by \textbf{7.5\%} and \textbf{15.8\%} under the third-person and wrist-only settings respectively. Unlike baselines that rely on instantaneous views, our consistent improvements across all tasks confirm that continuously integrating historical context into a persistent world state memory effectively overcomes wrist-view partial observability during robotic manipulation.

\begin{table*}[t]
  \centering
  \setlength{\tabcolsep}{12pt}
  \resizebox{\textwidth}{!}{
    \begin{tabular}{
      l|c|*{6}{c}|c
    }
      \toprule
      \multicolumn{1}{c|}{\multirow{2}{*}[-1.8ex]{\makecell[c]{Method}}}
        & \multicolumn{1}{c|}{\multirow{2}{*}[-1.8ex]{\makecell[c]{Cameras}}}
        & \multicolumn{7}{c}{\textbf{General Tasks}} \\
      \cmidrule(lr){3-9}
        & 
        & \makecell[c]{Pepper\\on Plate}
        & \makecell[c]{Pepper\\in Box}
        & \makecell[c]{Stack\\ Cubes}
        & \makecell[c]{Carrot\\ on Plate}
        & \makecell[c]{Cube\\in Drawer}
        & \makecell[c]{Can\\in Drawer}
        & \makecell[c]{Avg. } \\
      \midrule

        $\pi_0$~\scriptsize\citep{black2024pi_0}
        & 3rd + wrist & 68.0 & 60.0 & 62.0 & 74.0 & 66.0 & 70.0 & 66.7\\
        MemoryVLA~\scriptsize\citep{shi2025memoryvla}
        & 3rd  & \underline{72.0} & \underline{64.0} & \underline{66.0} & \underline{78.0} & \underline{70.0}  & \underline{74.0} & \underline{70.7}\\
        MemoryVLA~\scriptsize\citep{shi2025memoryvla}
        & wrist & 62.0 & 56.0 & 58.0 & 70.0 & 62.0 & 66.0 & 62.3\\
        {\textbf{AtlasVLA (Ours)}}
        & wrist & \textbf{78.0} & \textbf{72.0} & \textbf{76.0} & \textbf{84.0} & \textbf{82.0}  & \textbf{80.0} & \textbf{78.7}
        \\
      \bottomrule
    \end{tabular}
  }
  \vspace{-3mm}
    \caption{\textbf{Quantitative results on the real-world general tasks.} 
    We report average success rates (\%) over 50 trials for each task.}
    \vspace{-2mm}
    \label{tab:real1}
\end{table*}

\begin{table}[t]
  \centering
  \renewcommand{\arraystretch}{1.0}
  \resizebox{\linewidth}{!}{
    \begin{tabular}{
      l|*{4}{c}|c
    }
      \toprule
      \multicolumn{1}{c|}{\multirow{2}{*}[-1.8ex]{\makecell[c]{Method}}}
       & \multicolumn{5}{c}{\textbf{Long-horizon Tasks}} \\
      \cmidrule(lr){2-6}
        & \makecell[c]{Change\\Cubes}
        & \makecell[c]{Stack Cubes\\ Order}
        & \makecell[c]{Clean\\Desk}
        & \makecell[c]{Pick Place\\Order}
        & \makecell[c]{Avg. } \\
      \midrule

      $\pi_0$
        & 54 & 50 & 52 & 52 & 52.0 \\
      MemoryVLA
        & \underline{62} & \underline{58} & \underline{60} & \underline{62} & \underline{60.5} \\
       {\textbf{AtlasVLA (Ours)}} & \textbf{74} & \textbf{66} & \textbf{68} & \textbf{70} & \textbf{69.5} \\
      \bottomrule
    \end{tabular}
 }
  \vspace{-3mm}
    \caption{\textbf{Results on the real-world long-horizon tasks.} 
    We report average success rates (\%) over 50 trials for each task.}
    \vspace{-4mm}
    \label{tab:real2}
\end{table}

\subsection{Real-World Evaluation}

\ptitle{Implementation Setup.}
We evaluate AtlasVLA on a real-world robot platform under a strict wrist-only setting, where only a wrist-mounted camera is used without any additional third-person views. We construct two task suites, including 6 general manipulation tasks and 4 long-horizon tasks, covering sequential manipulation, rearrangement, and multi-stage interactions. All methods are evaluated over 50 trials per task using identical robot configurations and task instructions. 


\ptitle{Evaluation on General Tasks.}
We first evaluate AtlasVLA on general manipulation tasks. 
As shown in Table~\ref{tab:real1}, AtlasVLA achieves an average success rate of \textbf{78.7}\% using only a wrist camera, outperforming all baselines with richer visual inputs. 
It surpasses MemoryVLA by \textbf{8.0\%} and \textbf{16.4\%} under third-person and wrist-only settings, respectively. 
This demonstrates that persistent world state memory effectively compensates for limited wrist-camera visibility, enabling reliable manipulation without multi-camera perception.

\ptitle{Evaluation on Long-Horizon Tasks.}
We further evaluate AtlasVLA on challenging long-horizon manipulation tasks. As reported in Table~\ref{tab:real2}, AtlasVLA achieves an average success rate of \textbf{69.5\%}, outperforming $\pi_0$ and MemoryVLA by \textbf{17.5\%} and \textbf{9.0\%}, respectively. The significant improvement highlights the importance of persistent spatio-temporal reasoning for long-horizon manipulation. While MemoryVLA improves over $\pi_0$ by leveraging temporal memory, it still struggles to maintain sufficient spatial context in multi-stage tasks. In contrast, AtlasVLA jointly models spatial and temporal information through its world latent memory, enabling the robot to preserve intermediate states, recover previously observed information, and execute complex manipulation sequences more reliably. These results validate that explicit world state memory is essential for scaling VLA models from reactive control toward robust long-horizon manipulation.

\begin{table}[htbp]
    \setlength{\tabcolsep}{8pt}
    \renewcommand{\arraystretch}{0.8}
    
    \centering
    \scriptsize
    \resizebox{1.0\linewidth}{!}{
    \begin{tabular}{l|l|cc}
    \toprule
    No.  &Methods        & LIBERO  & Real-world Long \\
    \midrule
    1)    &w/o World State Memory  &{93.5}  &{54.0} \\
    2)    &w/o Ego-Working Memory  &{95.0}  &{56.5} \\
    {3)}         & AtlasVLA &\textbf{97.6}  &\textbf{69.5} \\
    \midrule
    4)         &w/o World State Update   &{94.6}  &{58.0} \\
    {5)}         &w World State Update  &\textbf{97.6}  &\textbf{69.5} \\
    \midrule
    6)         & w/o Spatial PE  &{96.4}  &{67.5} \\
    7)         &w/o Temporal PE   &{96.8}  &{65.0} \\
    {8)}         &Spatio-Temporal PE&\textbf{97.6}  &\textbf{69.5} \\
    \midrule
    9)         &w/o World State Conditioning   &   {95.2}  &{61.5} \\
    {10)}         &w World State Conditioning   &\textbf{97.6}  &\textbf{69.5} \\
    
    \bottomrule
    \end{tabular}
    }
    \vspace{-3mm}
    \caption{Ablation study on LIBERO and real-world tasks. }
    \label{Ablation}
    \vspace{-4mm}
\end{table}

    
    

\subsection{Ablation Studies}
\label{subsec:ablation}

We conduct systematic ablation studies on LIBERO and real-world long-horizon tasks (Table \ref{Ablation}) to evaluate each component of our architecture under the strict wrist-only constraint.

\ptitle{Core Memory Modules.} 
Discarding world state memory (row 1) triggers a catastrophic performance collapse in real-world tasks (69.5\% $\rightarrow$ 54.0\%), compellingly demonstrating that our spatial map memory is indispensable for enabling robust robotic manipulation relying solely on a wrist camera.
Similarly, removing ego-working memory (row 2) causes a 13.0\% drop, proving that latent ego-working memory is imperative for preventing intention drift on long-horizon tasks.

\ptitle{Memory Update Strategy.} 
Replacing our spatio-temporal update with naive memory accumulation (row 4) degrades real-world success by 11.5\%, verifying the effectiveness of our TSDF-inspired voxel aggregation and sliding window.

\ptitle{Positional Embedding.} 
Removing spatial (row 6) or temporal PE (row 7) drops real-world performance by 2.0\% and 4.5\%, respectively, confirming the benefit of 4D positional embeddings for spatio-temporal reasoning.

\ptitle{World State Conditioning.} 
Removing world state attention  (row 9) reduces real-world success by 8.0\% confirming that injecting comprehensive world state in action generation is indispensable for synthesizing precise actions.

\section{Conclusion}
We present AtlasVLA, a novel Vision-Language-Action (VLA) framework that addresses the lack of a persistent world state in current reactive models, enabling robust wrist-only manipulation. To overcome severe partial observability, AtlasVLA introduces a Persistent World Memory featuring a voxel-hashed spatial map to preserve global world state, alongside an Ego-Working Memory to track task progression and prevent intention drift. Ultimately, the action output is achieved through the world-conditioned DiT, AtlasVLA achieves state-of-the-art performance across simulation and real-world benchmarks. Remarkably, using solely a wrist camera, it outperforms comprehensive multi-view baselines and boosts long-horizon success by +9.4\% on LIBERO-Long and +17.5\% in real-world tasks, establishing a scalable paradigm for persistent world state modeling.

\bibliography{aaai2027}


\clearpage
\appendix

\section{Implementation Details}

\subsection{AtlasVLA: Architecture}
In this section, we provide a comprehensive breakdown of the internal architectural components of AtlasVLA. Apart from our persistent world-ego state memory, our framework is fundamentally decoupled into three interconnected modules: a multimodal visual encoder for egocentric perception, a large language model for semantic reasoning, and a step-wise conditioned Diffusion Transformer (DiT) for action generation.

\ptitle{Visual Encoder.} 
To process the raw egocentric sensory inputs, AtlasVLA employs a dual-stream perception module. The primary RGB stream utilizes frozen vision encoders (DINOv2~\cite{oquab2024dinov2} and SigLIP~\cite{zhai2023sigmoid}) to extract 2D visual tokens directly from the wrist-camera observations. Concurrently, to construct the persistent 4D world representation, the spatial stream leverages a fine-tuned Depth Anything v3~\cite{lin2025depth} to provide robust streaming depth estimation. These depth maps are dynamically converted alongside the current camera extrinsics to facilitate strict spatial back-projection. 

\ptitle{Large Language Model.} 
The core cognitive and reasoning engine of AtlasVLA is instantiated as a LLaMA-2 7B decoder-only Large Language Model~\cite{touvron2023llama2}. We use OpenVLA-7b~\cite{kim2024openvla} as our pre-training model, and the LLM processes a multimodal sequence comprising the text instruction tokens, visual tokens, world state tokens and a dedicated set of learnable intent queries. Rather than directly decoding low-level actions, the LLM functions as a high-level intent tracker and semantic router. By executing cross-attention mechanisms over the historical context and current visual observations, it outputs compact ego-working latent states. These states are subsequently deposited into the ego-working memory bank, ensuring the agent maintains a coherent tracking of task progress and ego state.

\ptitle{Action Expert.} 
The translation of abstract world-ego states into continuous robotic control is governed by an action expert parameterized $\sim 300$M weights. Inspired by the highly expressive generative capabilities of CogACT~\cite{li2024cogact}, we adopt a Diffusion Transformer (DiT) architecture~\cite{peebles2023scalable}, which we fundamentally augment to support our dual-memory formulation. To effectively ground the action generation in the physical environment, we discard standard global conditioning in favor of a decoupled, step-wise dual-attention mechanism. At each diffusion step, the noisy action tokens sequentially attend to the retrieved historical ego-working context via an ego-working attention layer, and to the comprehensive scene representation via a world state attention layer. 
During online inference, the action sequence is formulated as a denoising process. To achieve high-frequency control, action decoding is accelerated using the Denoising Diffusion Implicit Models (DDIM)~\cite{song2020denoising} scheduler over 10 denoising steps. To further enhance the robustness of the synthesized trajectories, we apply Classifier-Free Guidance (CFG)~\cite{ho2022classifier} with a scale of 1.5. 

\subsection{Additional Training Details}

\ptitle{Loss Function.} 
Following~\cite{li2024cogact}, we model the continuous action synthesis as a denoising process parameterized by our step-wise conditioned Diffusion Transformer (DiT). The primary optimization objective is to train the action expert to accurately reconstruct the ground-truth action chunks $A_t$ from Gaussian noise, strictly conditioned on our proposed dual-state representations.

In the forward diffusion process, we iteratively add Gaussian noise $\epsilon \sim \mathcal{N}(0, \mathbf{I})$ to the ground-truth action chunk $A_t$ over $K$ steps to obtain the noisy action $a_k$ at step \begin{equation}
a_k = \sqrt{\bar{\alpha}_k} A_t + \sqrt{1 - \bar{\alpha}_k} \epsilon
\end{equation}
where $\bar{\alpha}_k$ follows a predefined cosine noise schedule. 

During the reverse process, the DiT network $\epsilon_{\theta}$ is trained to predict the injected noise $\epsilon$. Crucially, departing from standard global conditioning pipelines, our network incorporates a decoupled, two-step conditioning mechanism. The noise prediction is sequentially modulated by the Ego-Working State context $C_t^{ego}$ (capturing temporal task progress) and the Persistent World State context $C_t^{world}$ (providing geometric and spatial grounding). The denoising objective is thus formulated as a Mean Squared Error (MSE) loss:
\begin{equation}
\mathcal{L}_{act} = \mathbb{E}_{A_t, \epsilon, k} \left[ \left\| \epsilon - \epsilon_{\theta}(a_k, k, C_t^{ego}, C_t^{world}) \right\|_{2}^{2} \right]
\end{equation}

To effectively enable Classifier-Free Guidance (CFG) during inference, we employ a condition dropout strategy during training. Specifically, the conditioning contexts $(C_t^{ego}, C_t^{world})$ are independently replaced with a learnable unconditional null token $\emptyset$ with a probability of $p_{drop} = 0.1$. This joint training mechanism forces the model to learn both conditionally and unconditionally, facilitating highly robust action execution under the guided sampling framework.

\begin{table}[htbp]
    \centering
    \resizebox{1.0\linewidth}{!}{
    \begin{tabular}{lc}
        \toprule
        \textbf{Hyperparameter} & \textbf{Value} \\
        \midrule
        Global batch size & 256 (32 $\times$ 8) \\
        Learning rate & $2 \times 10^{-5}$ \\
        Action chunk size & 16 \\
        Inference denoising steps & 10 \\
        CFG scale (classifier-free guidance) & 1.5 \\
        Condition dropout & 0.1 \\
        Capacity of world state memory & 2048\\
        The voxel size of the world state & 0.025m \\
        Number of ego-working memory retrieval layer &2\\
        Number of ego-guided world retrieval layer &4 \\
        Number of ego-working tokens &4 \\
        \bottomrule
    \end{tabular}
    }
    \caption{Training and model hyperparameters for AtlasVLA.}
    \label{tab:hyperparameters}
\end{table}

\ptitle{Hyperparameter Setting.}
The detailed training and architectural hyperparameters for AtlasVLA are summarized in Table~\ref{tab:hyperparameters}. For the model optimization, we utilize a global batch size of 256 (distributed as 32 samples across 8 GPUs) and set the learning rate to $2 \times 10^{-5}$. Regarding our proposed dual-memory architecture, the Persistent World State Memory maintains a maximum capacity of 2048 tokens and updates the 3D environment using a spatial voxel resolution of $0.025\text{m}$. Concurrently, the Ego-Working Memory tracks historical task progress using 4 compact ego-working tokens. To effectively integrate these contextual representations into the action expert, we employ 2 cross-attention layers for the ego-working memory retrieval and 4 layers for the ego-guided world retrieval. For the continuous control synthesis, the model predicts trajectories with an action chunk size of 16. During online inference, the denoising process is executed over 10 steps. Finally, to ensure robust action alignment with the high-level semantic intent, we apply Classifier-Free Guidance (CFG) with a scale of 1.5.

\section{Additional Experimental Results}

\subsection{Additional Simulation Details}

\ptitle{LIBERO.}
The LIBERO benchmark~\cite{liu2023libero} serves as a comprehensive simulation environment to systematically assess the compositional generalization and sequential decision-making capabilities of embodied agents across five distinct task suites: \textit{LIBERO-Spatial}, \textit{LIBERO-Object}, \textit{LIBERO-Goal}, \textit{LIBERO-Long}, and \textit{LIBERO-90}~\cite{liu2023libero}. Because these tasks require accomplishing a series of coherent sub-goals under diverse initial conditions, they pose a severe challenge for standard memoryless models, particularly in the \textit{Long} and \textit{90} suites where compounding errors from spatial blind spots and intention drift frequently lead to catastrophic failures. For our experimental protocol, we strictly utilize 50 expert demonstrations per task, training individual policies for the Spatial, Object, and Goal suites for 20,000 gradient steps each, while adopting a unified co-training strategy for the Long and LIBERO-90 suites over 40,000 steps~\cite{shi2025memoryvla}. During evaluation, we report the mean success rate computed over 50 independent execution rollouts for each task, and to ensure rigorous evaluation without validation bias, all quantitative metrics are extracted directly from the final training checkpoint. Crucially, to validate our dual-memory architecture under partial observability, our LIBERO experiments rely strictly on a single wrist-mounted camera, discarding the third-person global views used by standard baselines.

\ptitle{RLBench.}
The RLBench framework~\cite{james2020rlbench} provides a geometrically complex simulation environment featuring multi-stage manipulation tasks with significant spatial and trajectory variations. To evaluate our model within this framework, we utilize 100 expert demonstrations per task and train the policy for 80,000 gradient steps. Despite the high geometric complexity that typically necessitates multi-view observations, we enforce a rigorous wrist-only perception constraint by exclusively utilizing a single $128 \times 128$ RGB egocentric view while entirely discarding the standard front, left, and right third-person cameras. Following the evaluation protocol established in recent literature~\cite{lou2026mask}, we assess the agent's performance across six representative tasks, conducting 20 independent execution trials per task to compute the average success rate. Consistent with our evaluation methodology on LIBERO, all quantitative results on the RLBench benchmark are derived directly from the final training checkpoint to ensure robust performance measurement and eliminate validation bias.

\subsection{Additional Real-world Settings}

\begin{figure}[t]
    \centering{\includegraphics[width=1.0\linewidth]{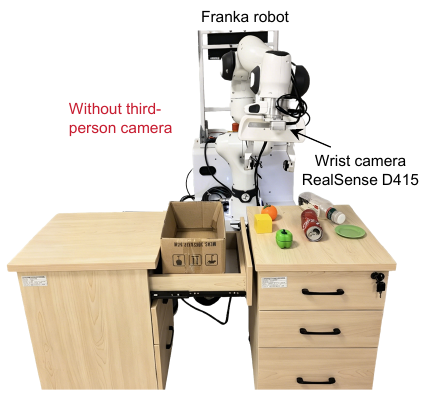}}  
    \caption{
    Our real-world robotic platform.
    }
    \label{fig:robot_setup}
\end{figure}

\ptitle{Hardware Setup.}
All real-world evaluations are conducted using a 7-DoF Franka robotic manipulator. Departing from conventional multi-camera configurations, our system strictly enforces a wrist-only observation constraint by exclusively utilizing a single Intel RealSense D415 camera mounted on the end-effector for egocentric perception. Any third-person global cameras are deliberately disabled and disconnected from the data flow to rigorously test the agent's spatial reasoning under severe partial observability. The physical workspace features an unstructured tabletop environment equipped with cabinets, sliding drawers, and a diverse set of everyday objects scattered across the manipulation area to support long-horizon tasks. All hardware components are synchronized and operated through the Robot Operating System (ROS), facilitating real-time sensorimotor control and action execution. The detailed visualization of our real-world robotic platform and the restrictive observation constraints are depicted in Figure~\ref{fig:robot_setup}.

\begin{table*}[t]
\centering
\begin{tabular}{lcccc}
\toprule
Methods &   Latency (s) & Throughput (Hz)   & GPU Memory (GB) & Success rate (\%)\\
\midrule
MemoryVLA~\cite{shi2025memoryvla} & 0.146 & 109.5 & 16.7 GB & 62.3\\
\textbf{AtlasVLA (Ours)} & 0.158 & 101.3 & 18.1 GB & 78.7\\
\bottomrule
\end{tabular}
\caption{Quantitative analysis of runtime efficiency and memory footprint. The inference latency and action generation throughput are measured to evaluate the feasibility of real-time robotic deployment.}
\label{tab:runtime}
\end{table*}

\ptitle{Task Design.}
We construct two distinct real-world task suites to evaluate the robustness and temporal reasoning capabilities of our framework under the strict wrist-only observation constraint. The first suite comprises 6 general manipulation tasks that focus on fundamental object interactions and spatial arrangements within the tabletop workspace. Furthermore, to rigorously assess the agent's ability to maintain a persistent world-ego state and mitigate historical forgetting, the second suite introduces 4 challenging long-horizon tasks. These extended scenarios cover complex sequential manipulation, scene rearrangement, and multi-stage interactions, requiring the policy to seamlessly overcome spatial blind spots and intention drift. In total, our real-world benchmark consists of 10 unique tasks, with all baseline comparisons systematically evaluated over 50 independent trials per task using identical robot configurations.

\ptitle{Evaluation Protocol.}
During the real-world inference phase, the performance on each task is measured across 50 independent execution trials, featuring randomized initial object placements and varied spatial layouts. A manipulation rollout is recorded as successful only if the robotic agent sequentially completes all required sub-goals and reaches the specified final target state within a designated step limit. For the demanding long-horizon suite, the evaluation specifically stresses the model's resilience against severe partial observability, requiring the policy to navigate dynamic perspective changes and extended temporal executions without catastrophic spatial or intention forgetting. To guarantee a rigorous and fair comparison, all evaluated baseline methods are deployed on the identical physical workspace.


\subsection{Runtime Analysis}\label{Runtime}
To evaluate the computational efficiency of our proposed dual-memory architecture for real-world robotic deployment, we conduct a comprehensive runtime analysis comparing AtlasVLA against the strong baseline, MemoryVLA~\cite{shi2025memoryvla}. As detailed in Table~\ref{tab:runtime}, despite the architectural integration of the continuous 4D Persistent World State updating and the step-wise Ego-Working State retrieval mechanisms, AtlasVLA maintains highly competitive inference efficiency. Specifically, the single-step inference latency experiences a marginal increase of only $0.012\text{s}$ (from $0.146\text{s}$ to $0.158\text{s}$) compared to the baseline, seamlessly sustaining a robust throughput of $101.3\text{Hz}$ that comprehensively exceeds the standard execution frequency required for real-time, closed-loop continuous control in physical robots. Furthermore, the hardware resource analysis reveals that AtlasVLA consumes $18.1\text{GB}$ of GPU VRAM, introducing a modest $1.4\text{GB}$ overhead over MemoryVLA. This indicates that our redundancy-aware latent consolidation strategy successfully bounds unbounded memory growth, ensuring the entire framework can still be efficiently deployed on a single consumer-grade GPU. Ultimately, when juxtaposed with the substantial $16.4\%$ absolute improvement in the real-world success rate, these minimal computational trade-offs firmly establish AtlasVLA as an exceptionally efficient and high-performing framework.

\subsection{Detailed Ablation Study}

\begin{table}[htbp]
    \setlength{\tabcolsep}{8pt}
    \renewcommand{\arraystretch}{0.8}
    
    \centering
    \scriptsize
    \resizebox{1.0\linewidth}{!}{
    \begin{tabular}{l|l|cc}
    \toprule
    No.  &Memory length        & LIBERO  & Real-world Long \\
    \midrule
    1)    &8  &{97.3}  &{66.4} \\
    2)    & \textbf{16 (Ours)} &{97.6}  &{69.5} \\
    {3)}  &32  &{97.2}  &\textbf{69.8} \\
    \bottomrule
    \end{tabular}
    }
    \caption{Ablation study on the memory length of the ego-working state across LIBERO and real-world tasks.}
    \label{Ablation1}
\end{table}

\ptitle{Memory Length of Ego-Working State.}
As shown in Table~\ref{Ablation1}, we ablate the memory length of the Ego-Working State to evaluate its impact on sequential decision-making. Setting the length to 16 yields the optimal balance, outperforming the shorter length of 8 on both LIBERO (97.6\% vs. 97.3\%) and real-world long-horizon tasks (69.5\% vs. 66.4\%). This performance gain confirms that an adequate temporal horizon is crucial for tracking task progress and mitigating intention drift. However, further expanding the memory to 32 provides marginal real-world benefits (69.8\%) while slightly degrading LIBERO performance (97.2\%). This indicates that excessively long memory introduces redundant historical noise, which dilutes the cross-attention mechanism without yielding meaningful gains. Consequently, we adopt a memory length of 16 as the optimal hyperparameter.

\begin{table}[htbp]
    \setlength{\tabcolsep}{8pt}
    \renewcommand{\arraystretch}{0.8}
    
    \centering
    \scriptsize
    \resizebox{1.0\linewidth}{!}{
    \begin{tabular}{l|l|cc}
    \toprule
    No.  & Voxel size       & LIBERO  & Real-world Long \\
    \midrule
    1)    &0.01  &{96.3}  &{65.7} \\
    2)    &\textbf{0.025 (Ours)}  &\textbf{97.6}  &\textbf{69.5} \\
    {3)}  &0.05 &{97.2}  &{64.0} \\
    {4)}  &0.1 &{95.9}  &{58.5} \\
    \bottomrule
    \end{tabular}
    }
    \caption{Ablation study on the voxel size of the world state across LIBERO and real-world tasks.}
    \label{Ablation2}
\end{table}

\ptitle{Voxel Size of World State.}
Table~\ref{Ablation2} details the ablation on the spatial voxel resolution used for constructing the Persistent World State. A voxel size of $0.025\text{m}$ yields the best performance, achieving the highest success rates on both LIBERO (97.6\%) and real-world long-horizon tasks (69.5\%). Decreasing the voxel size to $0.01\text{m}$ degrades performance (96.3\% and 65.7\%, respectively), likely because overly fine-grained voxelization results in excessive spatial sparsity and introduces representation noise, thereby affecting the update of the world state. Conversely, increasing the voxel size to $0.05\text{m}$ and $0.1\text{m}$ causes a severe performance drop, particularly in complex real-world scenarios (falling to 64.0\% and 58.5\%), as coarse grids over-compress essential geometric structures and discard the fine spatial details required for precise manipulation. Therefore, the $0.025\text{m}$ resolution provides the optimal balance between geometric fidelity and representational compactness.

\subsection{Qualitative Results}

\begin{figure*}[t]
    \centering{\includegraphics[width=1.0\textwidth]{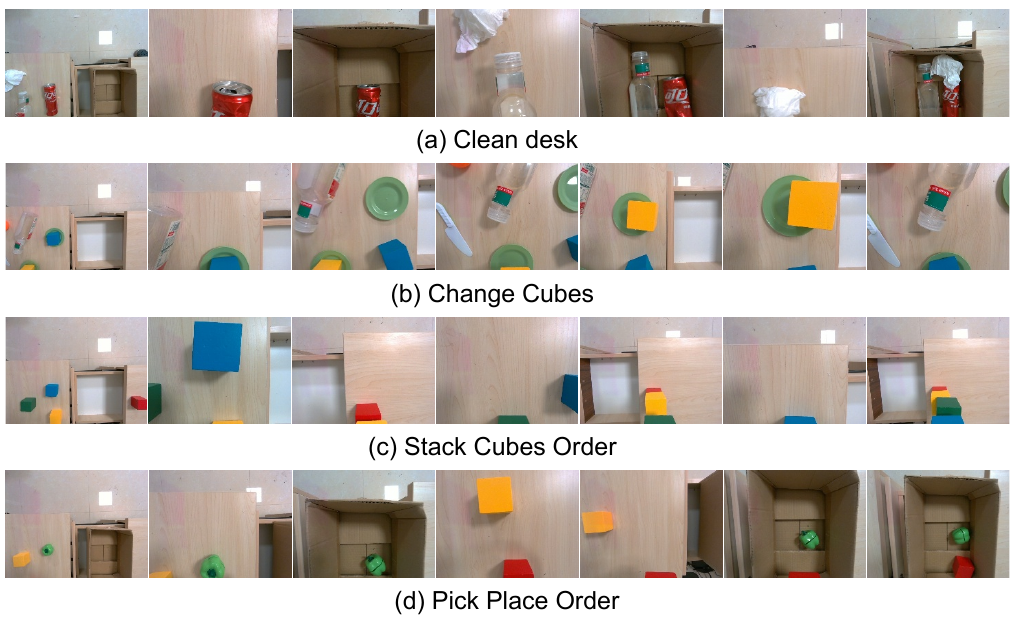}}  
    \caption{
    Qualitative results of AtlasVLA on real-world long-horizon tasks.
    }
    \label{fig:Qualitative1}
\end{figure*}

\begin{figure*}[t]
    \centering{\includegraphics[width=1.0\textwidth]{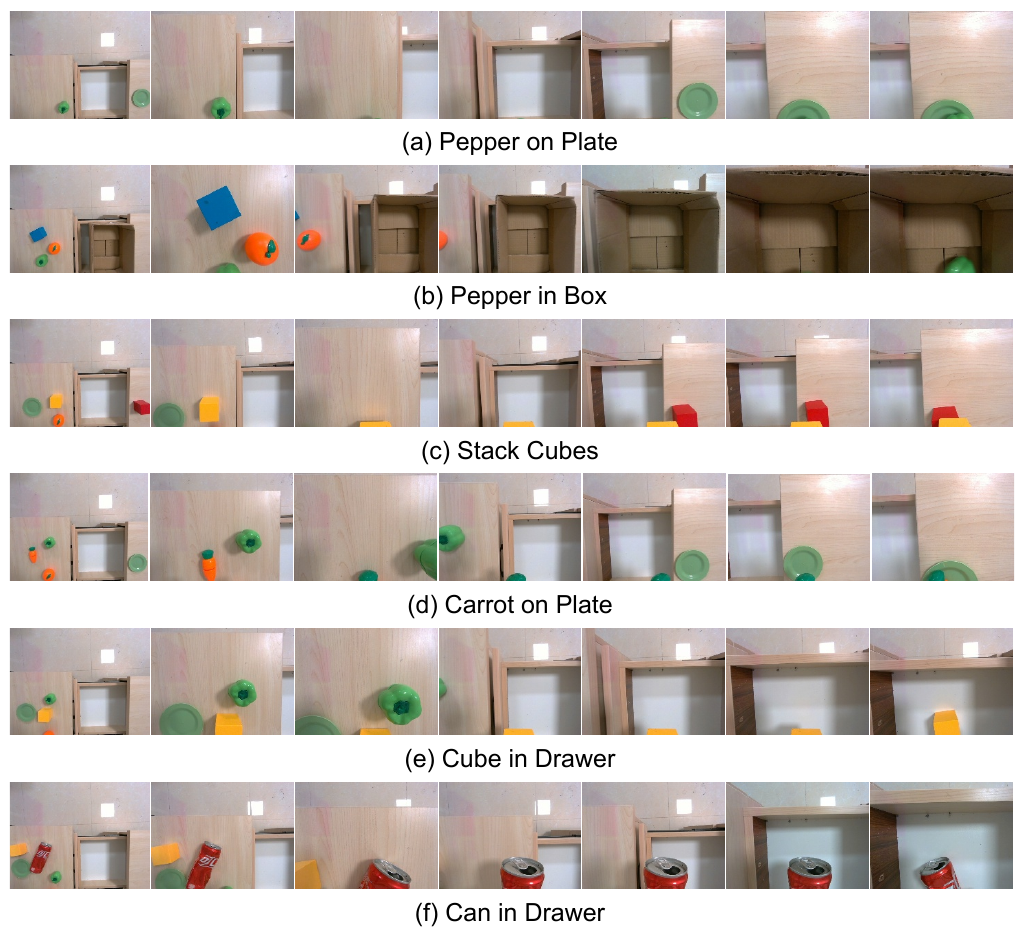}}  
    \caption{
    Qualitative results of AtlasVLA on real-world general tasks.
    }
    \label{fig:Qualitative2}
\end{figure*}

\ptitle{Qualitative Results in Real-world.}
To provide a more intuitive understanding of our method's performance under strict wrist-only partial observability, we present step-by-step visual execution trajectories of AtlasVLA deployed in the physical environment. Figure~\ref{fig:Qualitative2} illustrates the qualitative results across the real-world general tasks, such as ``Pepper in Box'' and ``Can in Drawer'', where the robotic agent is tasked with fundamental pick-and-place and spatial rearrangement operations. Due to the narrow field-of-view inherent to the wrist-mounted camera, the target receptacle inevitably exits the visual frame as the end-effector approaches the initial object to be grasped; however, the execution sequences clearly demonstrate that AtlasVLA successfully manipulates the targets without losing spatial consistency, directly validating the effectiveness of the proposed 4D Persistent World State memory in continuously integrating transient local observations to retain the geometric workspace layout. Furthermore, Figure~\ref{fig:Qualitative1} visualizes the execution progress on the highly demanding real-world long-horizon tasks, including ``Clean desk'', ``Change Cubes'', ``Stack Cubes Order'', and ``Pick Place Order''. These extended scenarios require the robot to accurately sequence multiple manipulation sub-goals amidst drastic and continuous viewpoint shifts. As depicted in the multi-step trajectories, AtlasVLA maintains coherent temporal execution across the entire horizon without erroneously repeating previous actions or losing track of the final objective. By seamlessly retrieving historical task progress from the Ego-Working State memory and grounding it within the updated Persistent World State, our policy effectively mitigates both intention drift and perception forgetting, ultimately executing complex, multi-stage behaviors with remarkable robustness.

\begin{figure*}[t]
    \centering{\includegraphics[width=1.0\textwidth]{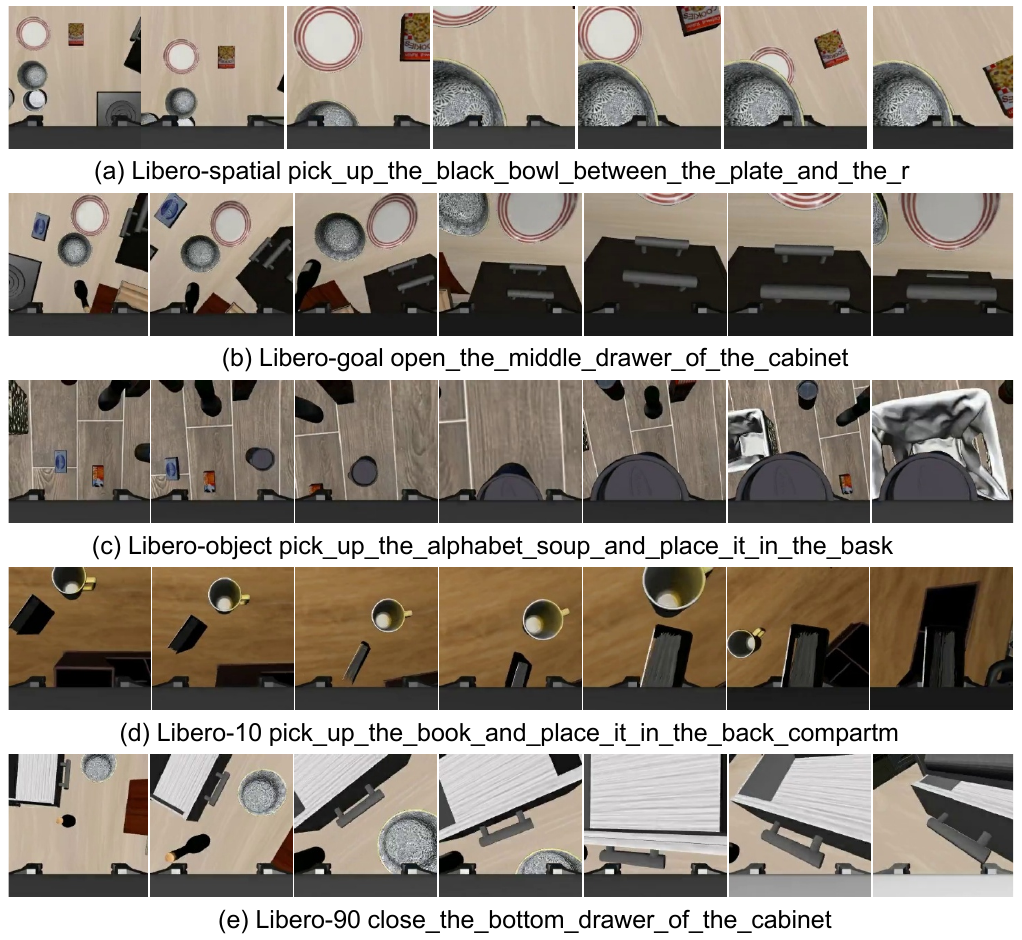}}  
    \caption{
    Qualitative results of AtlasVLA on LIBERO Benchmark.
    }
    \label{fig:Qualitative3}
\end{figure*}

\begin{figure*}[t]
    \centering{\includegraphics[width=1.0\textwidth]{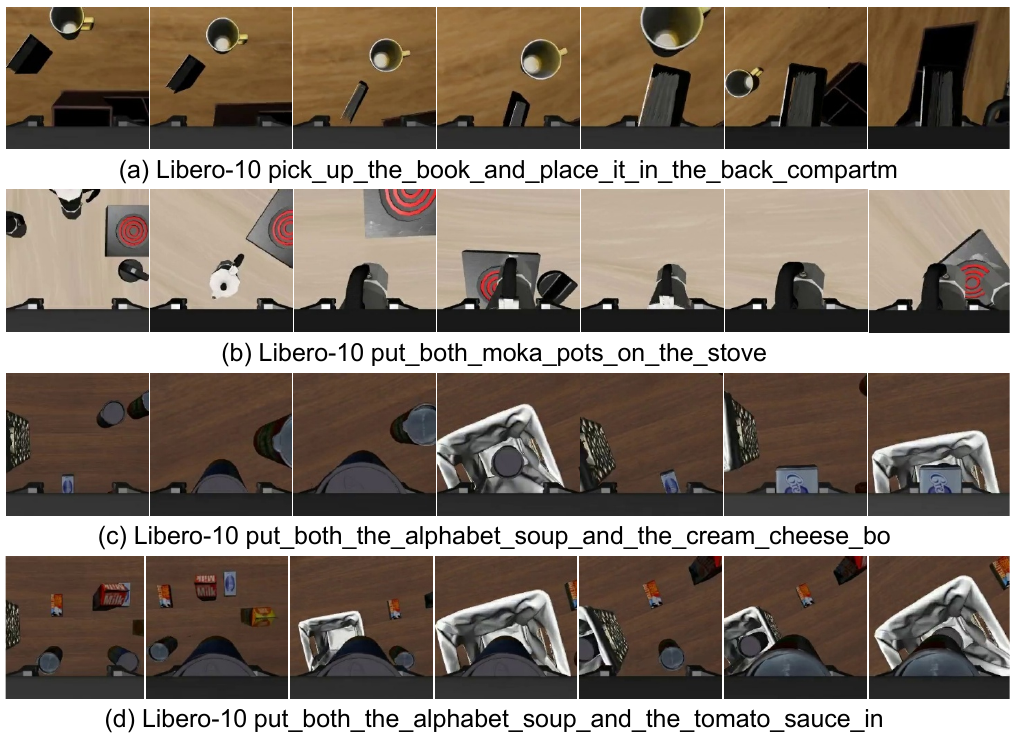}}  
    \caption{
    More Qualitative results of AtlasVLA on long LIBERO-10 Benchmark.
    }
    \label{fig:Qualitative4}
\end{figure*}

\ptitle{Qualitative Results on LIBERO.}
To further substantiate the efficacy of our proposed dual-memory architecture in simulated environments, we visualize the step-by-step execution trajectories of AtlasVLA across the comprehensive LIBERO benchmark suites. Figure~\ref{fig:Qualitative3} illustrates the agent's robust manipulation capabilities across diverse evaluation benchmarks, including the Spatial, Goal, Object, and 90 subsets. Despite the inherently restricted field-of-view of the wrist-mounted camera, the agent seamlessly handles varied spatial layouts, novel object instances, and diverse task instructions without catastrophic perception failures. For instance, during tasks requiring precise spatial alignment, the 4D Persistent World State memory effectively retains the global geometric layout of target receptacles—such as baskets or drawers—even as they momentarily exit the egocentric frame, facilitating accurate and collision-free interactions. Furthermore, Figure~\ref{fig:Qualitative4} details the qualitative performance on the demanding LIBERO-10 (Long) benchmark, which involves complex, multi-stage tasks such as sequentially placing multiple distinct items (e.g., moka pots or soup cans) into designated locations. In these extended manipulation horizons, the wrist perspective undergoes drastic and continuous shifts, a scenario that typically causes memoryless reactive baselines to suffer from severe intention drift. However, guided by the Ego-Working State memory, AtlasVLA successfully tracks its historical task progress, ensuring it implicitly remembers previously completed sub-goals and accurately transitions to subsequent actions to fulfill the long-horizon tasks.


\end{document}